\documentclass{article}
\usepackage[utf8]{inputenc}
\usepackage{iclr2022_conference,times}
\usepackage[colorlinks]{hyperref}
\setcitestyle{square}
\usepackage{graphicx}
\graphicspath{ {./plots/} }
\iclrfinalcopy

\title{Neighboring state-based Exploration\\ for Reinforcement Learning}

\author{%
  Yu-Teng Li \\ UC Berkeley \And
  Justin Lin \\ UC Berkeley \And
  Jeffery Cheng \\ UC Berkeley \And
  Pedro Pachuca \\ UC Berkeley
}

\date{December 2022}

\begin{document}

\maketitle

\vspace{-1em}
\begin{abstract}
  Reinforcement Learning is a powerful tool to model decision-making processes. However, it relies on an exploration-exploitation trade-off that remains an open challenge for many tasks. In this work, we study neighboring state-based, model-free exploration led by the intuition that, for an early-stage agent, considering actions derived from a bounded region of nearby states may lead to better actions when exploring. We propose two algorithms that choose exploratory actions based on a survey of nearby states, and find that one of our methods, ${\rho}$-explore, consistently outperforms the Double DQN baseline in an discrete environment by 49\% in terms of Eval Reward Return. Code available at \url{https://github.com/thekevinli/rho_exploration}.
\end{abstract} 

\section{Introduction}

A popular area of recent study in Reinforcement Learning (RL) is that of exploration methods. RL depends on an exploration-exploitation trade-off \citep{shen2022exploration} where an  agent must balance between taking an action prescribed by an existing policy or a novel action according to some other algorithm. Such an exploration algorithm attempts to deviate from a policy in a beneficial way as to find actions that efficiently maximize downstream returns but have not yet been encoded into the policy.  

Many such exploration methods have been proposed and the research community has found varying success across differing domains. Exploration algorithms and heuristics inject an intrinsic bias into RL algorithms that operate in environments with or without frequent, extrinsic feedback in the form of rewards. As a result, the effectiveness of exploration methods have been found to depend heavily on the domain studied. 

Some leading exploration methods include randomized exploration and structured exploration. Randomized explorations include $\epsilon$-greedy and Boltzmann exploration which introduce probabilistic choice among the optimal action according to the current policy and other, potentially entirely random, actions. Structured exploration methods include density estimation and entropy maximization methods that take advantage of the form of the problem to mathematically induce bias towards actions that lead to states maximally different than the current policy achieves \citep{Bellemare, EntropyMax}. 

\section{Rethinking Exploration}

We now look upon the intuition of the form of RL problems and combine structural and randomized exploration methods. 

When an RL agent is early in its training process, the agent generally relies more heavily on exploration as its policy does not yet encode many reasonable, much less optimal, state-action trajectories. With this in mind, especially in continuous environments or discrete environments with large state spaces, we may frequently encounter a state that our current policy does not have a good action for
yet is surrounded by nearby states that may have some good policy-defined action. Drawing on the idea that nearby states around some epicenter state are typically similar to the epicenter state, considering the actions taken by our policy in nearby states can help choose a good action on the current state \citep{stateSimilarity}. This is inspired by adversarial attacks which utilize guided perturbations \citep{bryniarskiadversarial}.

In most environments, we may only have access to observations of state, not fully-descriptive states. In these cases, we simply find nearby observations in observation space rather than state space.

There are some counterexamples to this proposition, such as in tasks where rewards are sparse, or any situation where high rewards are only earned by a small set of states or states far apart from each other \citep{LADOSZ20221}. Yet, traditional exploration methods also suffer dramatically in these scenarios, and therefore exploration in sparse reward settings is not the focus of this study.   

\subsection{Motivating Example}

A motivating example for our reasoning is that of Atari games. Such games generally have a large state space and a limited set of discrete actions \citep{Atari}. In these games, new states are reached frequently, and states near one another are very similar. When the policy reaches a new state, it may be beneficial to understand how the policy behaved in nearby states and, with some probability, consider those actions instead of the action prescribed by its policy.  

\section{Our approach}

To create an exploration schema that considers surrounding states, we take two approaches. In the following sections, we employ conventional notations in RL research, namely $\pi$ denotes the policy, $\pi^{*}$ denotes the optional policy, $r(s, a)$ for the reward, which depends on state $s$ and action $a$.

\subsection{Direct State Perturbation  ($\rho$-explore)}

\subsubsection{Algorithm}
$\rho$-explore, our first method, employs direct perturbation of our state. Fundamentally, the intuition of this method goes in two directions: if a state is good, nearby states should also be good; if a state is adversarially positive, or merely a local optimum, querying nearby states wisely can serve as an effective regularizer.

To ensure that nearby states are meaningful queries for \textit{local} exploration, we limit our consideration to states within some $\rho$ away from our current state. Formally, let $s$ be our current state and let $\rho$ be our perturbation bound. We employ this perturbation bound as a $p$-norm ball, most often taking the form of $||\cdot||_2$ or $||\cdot||_\infty$. For the sake of simplicity, we use the $l2$-norm for the remaining section. 

To find states within our region, we sample randomly. Concretely, we want to find state $s'$ such that $||s' - s||_2 < \rho$. Programatically, we simply generate random noise, $\delta$, with an $l2$-norm less than $\rho$ and let $s' = s + \delta$. We first generate $n$ new states $s'$ all within $\rho$ of the original state $s$. Then, we run $\lambda$ step mini-rollouts from each $s'$ according to our policy, where $\lambda$ is a hyperparameter, and score these $s'$ using the following function:
\begin{equation}
\label{eq:sample_score}
score(s) =[\ \sum_{i \in \lambda} r(s_i, a_i)\ ] + max_a\ Q(s_{\lambda+1}, a_{\lambda+1})
\end{equation}

Using these scores, we may subsequently determine the next action by $\pi(\ argmax_s [score(s)]\ )$, essentially picking the nearby state $s'$ that yields the highest score to determine the next action. Alternatively, the $argmax$ function can be replaced with the $mode$ of top $K$\% of the $n$ nearby states, ranked by their scores. We consider both cases in experiments, with the former variant named $max$ and the latter $mode$. 

\subsubsection{Implementation Details}
In practice, $\rho$-explore is applied along with $\epsilon$-greedy to encourage both local and global (uniformly random) exploration. We define a hyperparameter $\phi$ for the probability of the agent choosing $\rho$-explore instead of selecting action randomly. In our setting, $\phi$ is either 20 or 50 percent of $\epsilon$, which is scheduled to decrease over the course of training. 

Empirically, we observe that the agent performs better when allowed to apply $\rho$-explore for a consecutive interval, instead of ricocheting between local ($\rho$-explore) and global exploration at any time step, and thus we experimented with $10, 50, 100$ steps for the optimal $\rho$-explore interval.

A few technical details before diving into experiments: We adopted Double Q-learning (DDQN) as our model-free baseline \citep{Double-Q-Learning}, with a 3-layer MLP policy network, tested on Lunar-Lander environment from OpenAI gym \citep{openAiGym}. All presented results are averaged over 3 random seeds. Detailed hyperparameters are presented in Table \ref{tab:DQN_rho_HP}. Note that in some graphs $\lambda>1$ results are not included due to computation limitation (\ a $\lambda=10$ run could take 6+ hours to complete\ ). 

\subsubsection{Experiments}

\begin{figure}[!ht]
\includegraphics[width=11cm]{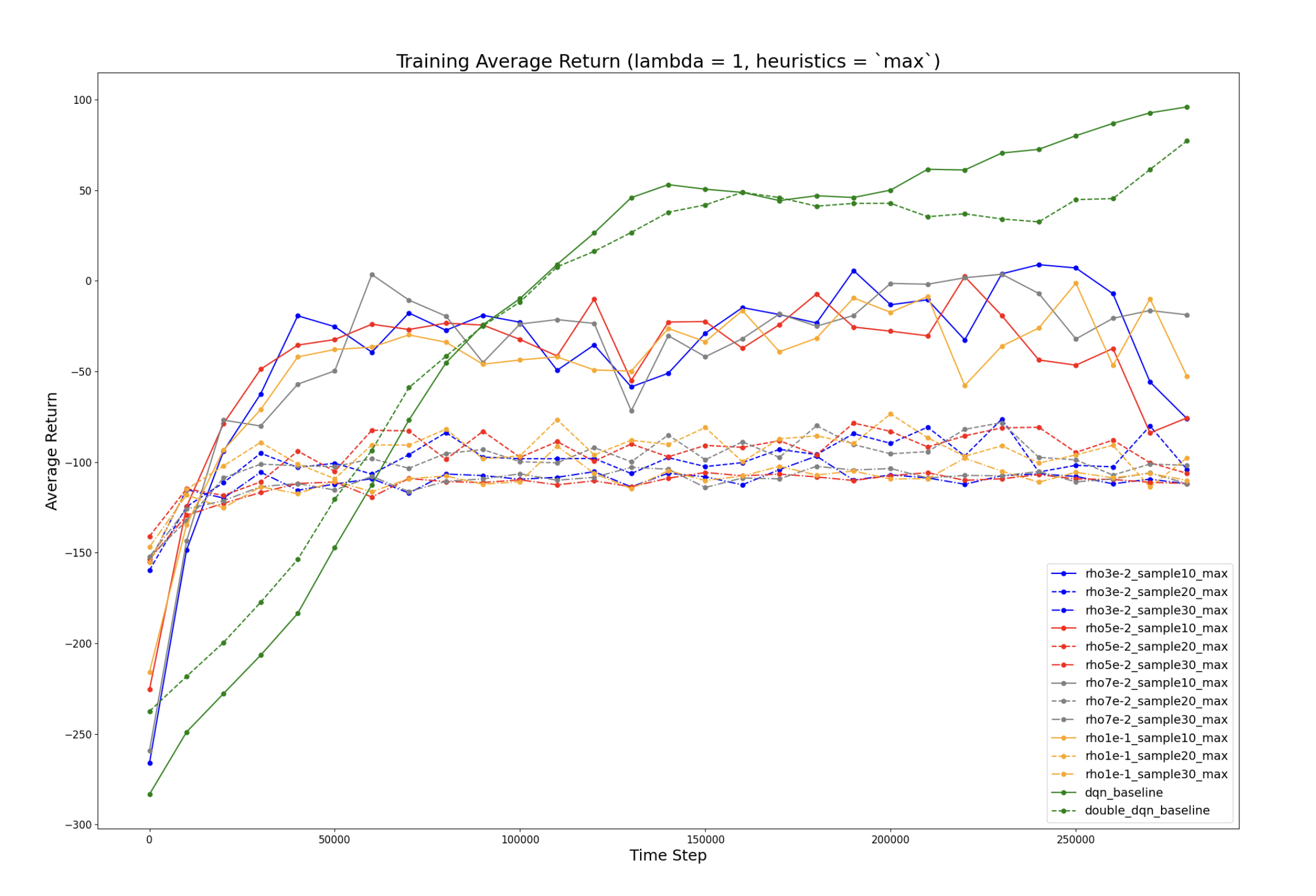}
\includegraphics[width=11cm]{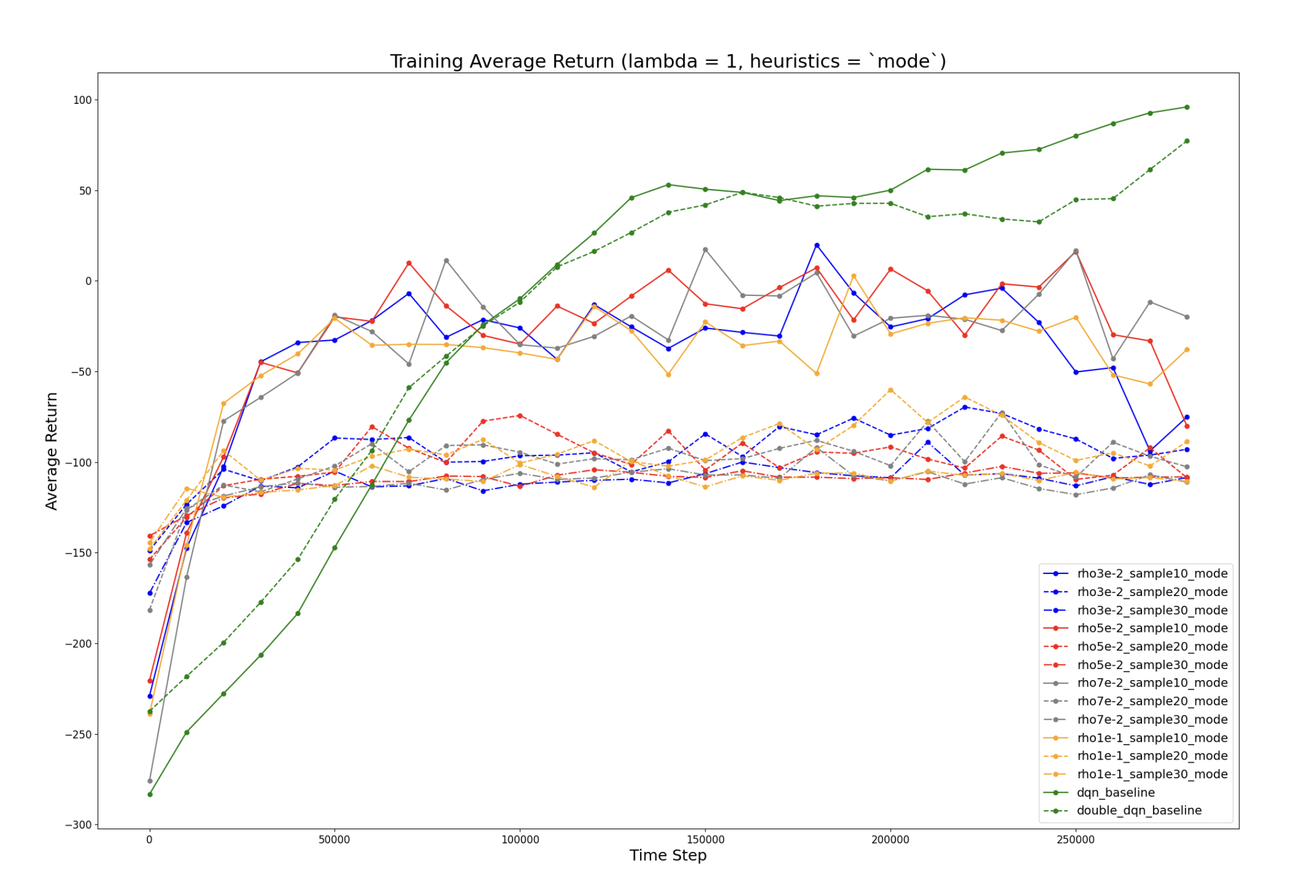}
\centering
\caption{Training Average Return with $\lambda=1$ step mini-rollouts; Upper: $max$, Lower: $mode$}
\label{fig:constant_ddqn_lambda1}
\end{figure}

\begin{figure}[!ht]
\includegraphics[width=12cm]{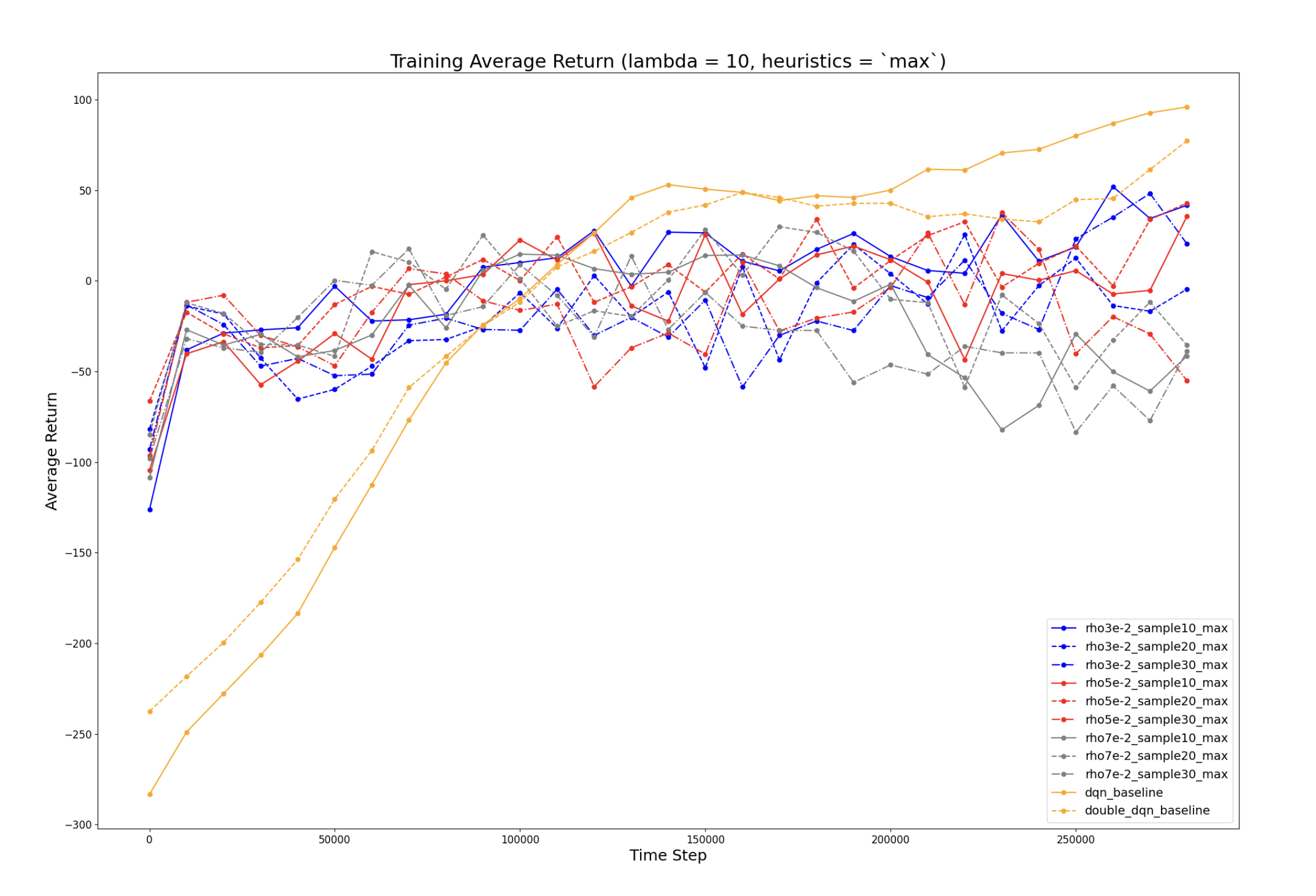}
\centering
\caption{Training Average Return with $\lambda=10$ step mini-rollouts}
\label{fig:constant_ddqn_lambda10}
\end{figure}

We started experimenting without the adaptive schedule described in section \textit{3.1.2}, meaning that $\rho$-explore is applied at an increasing likelihood over time (inversely correlated to $\epsilon$, which decays gradually) and capped at 50\%. Results are presented in Figure \ref{fig:constant_ddqn_lambda1} and \ref{fig:constant_ddqn_lambda10}. 

Both graphs for $max$ and $mode$ sampling heuristics indicate that while agents with $\rho$-explore began with a higher average reward than the vanilla Double-DQN baseline, ie. without our method, the reward function quickly plateaus out for agents with $\rho$-explore. Eventually, our method fails to outperform the vanilla baseline. 

We believe that such a phenomenon occurs due to not tuning down the exploration probability as the policy refines. While we hope that as the probability for $\epsilon$-greedy exploration gradually decreases $\rho$-explore can regularize the training by continuing to explore locally, the stochasticity induced by exploration hinders the performance as the policy improves. We thus redesigned $\rho$-explore with intermittent scheduling, and discovered a much better result. 

Additionally, since we observe that $max$ and $mode$ do not result in any nontrivial difference in reward return, for the remaining experiments, we will only be using $max$ setting. Both graphs also indicate that as the sample size $n$ increases, the plot becomes smoother but the final return decreases.

\begin{table}[!ht]
    \centering
    \begin{tabular}{c|c}
         $\rho$ & .03, .05, .07, .1  \\ \hline
         $n$ & 10, 20, 30 \\ \hline
         sampling heuristics & 'mode', 'max' \\ \hline
         $\rho$-explore period & 10, 50, 100 \\ \hline
         $\rho$-explore likelihood ($\phi$) & .2, .5 \\ \hline
         $\lambda^*$ & 1, 5, 10 \\
    \end{tabular}
    \caption{DQN $\rho$-explore hyper-parameters}
    \label{tab:DQN_rho_HP}
\end{table}

\begin{figure}[ht!]
\includegraphics[width=11cm]{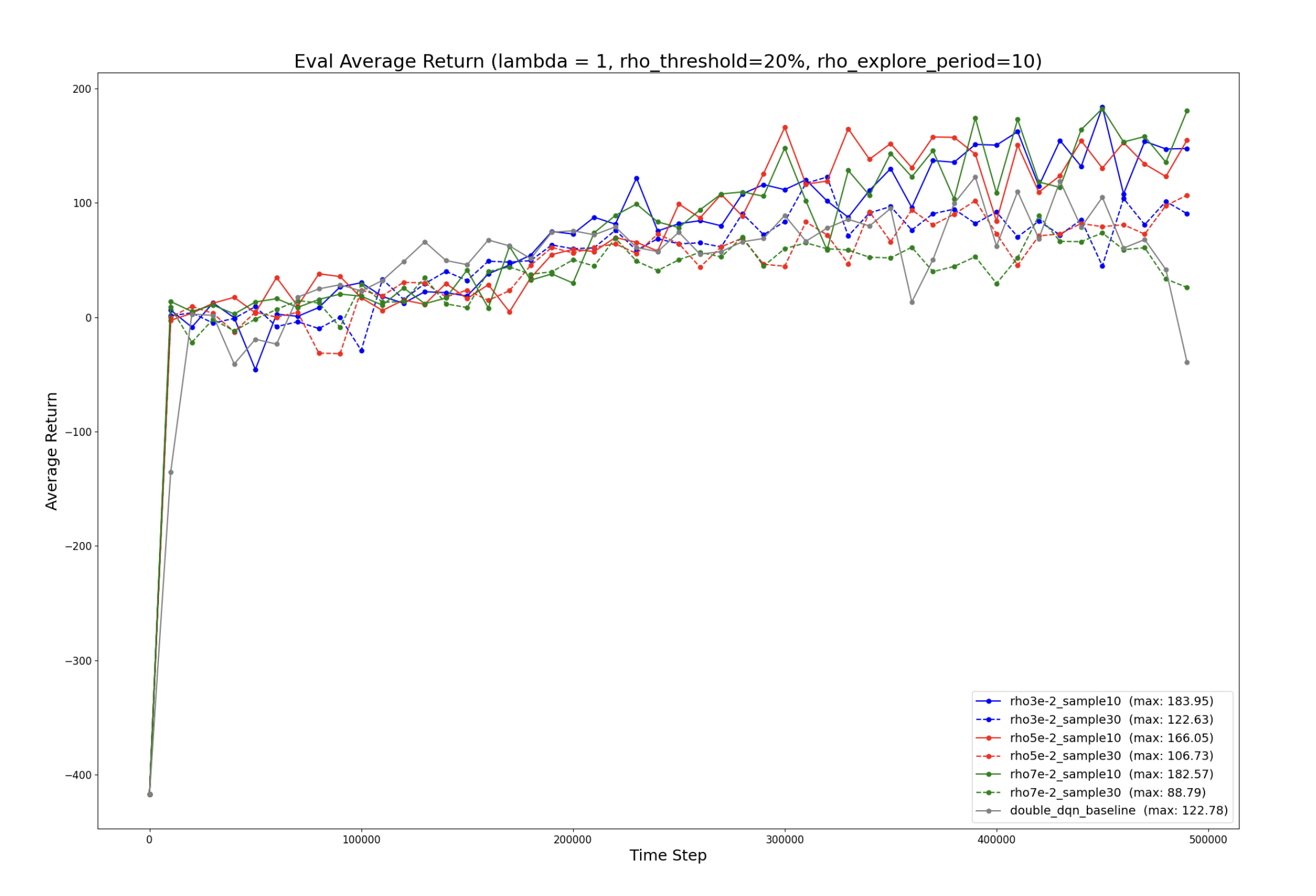}
\includegraphics[width=11cm]{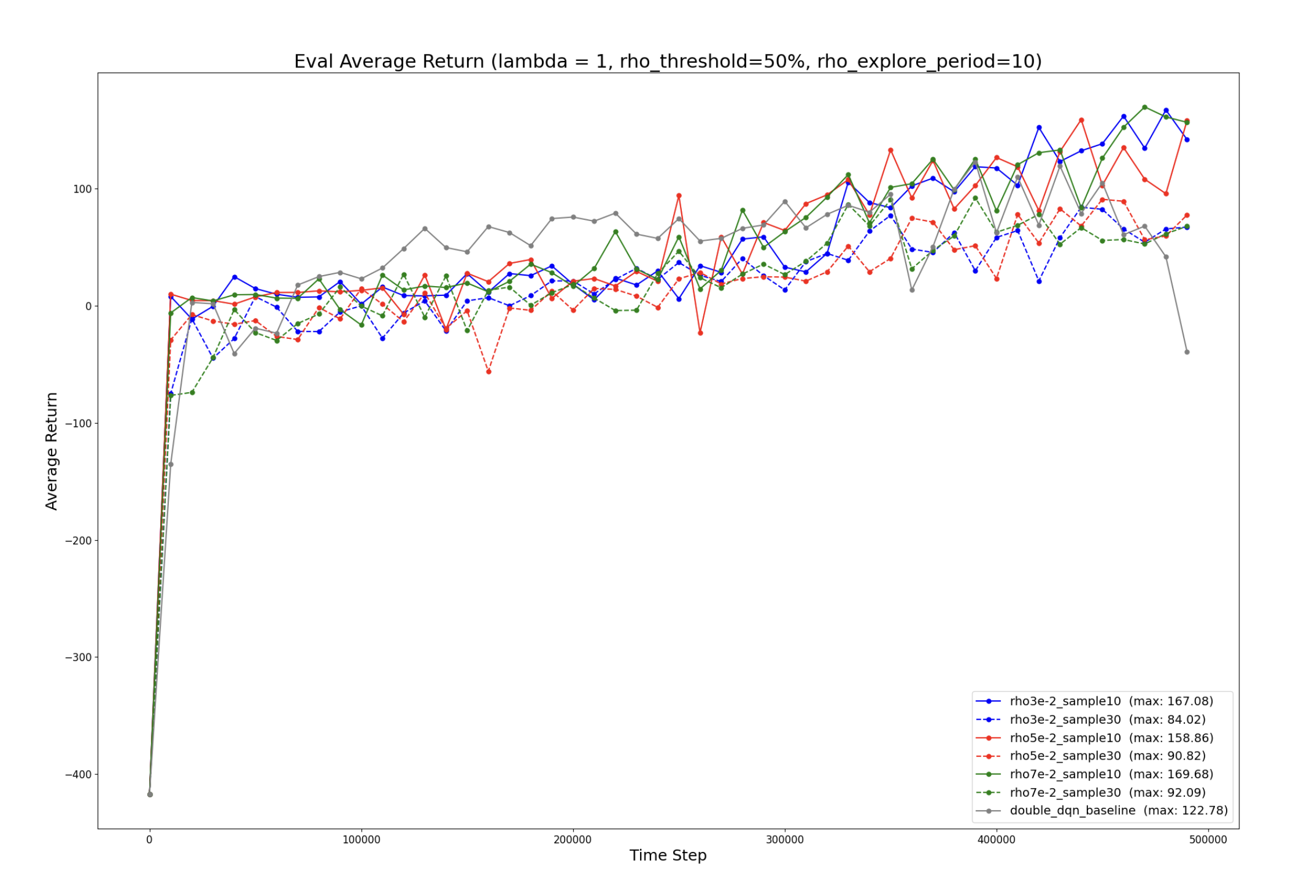}
\centering
\caption{Eval Average Return with $\lambda=1$ step; $\rho$-explore interval=10}
\label{fig:intermittent_interval10}
\end{figure}

\begin{figure}[ht!]
\includegraphics[width=11cm]{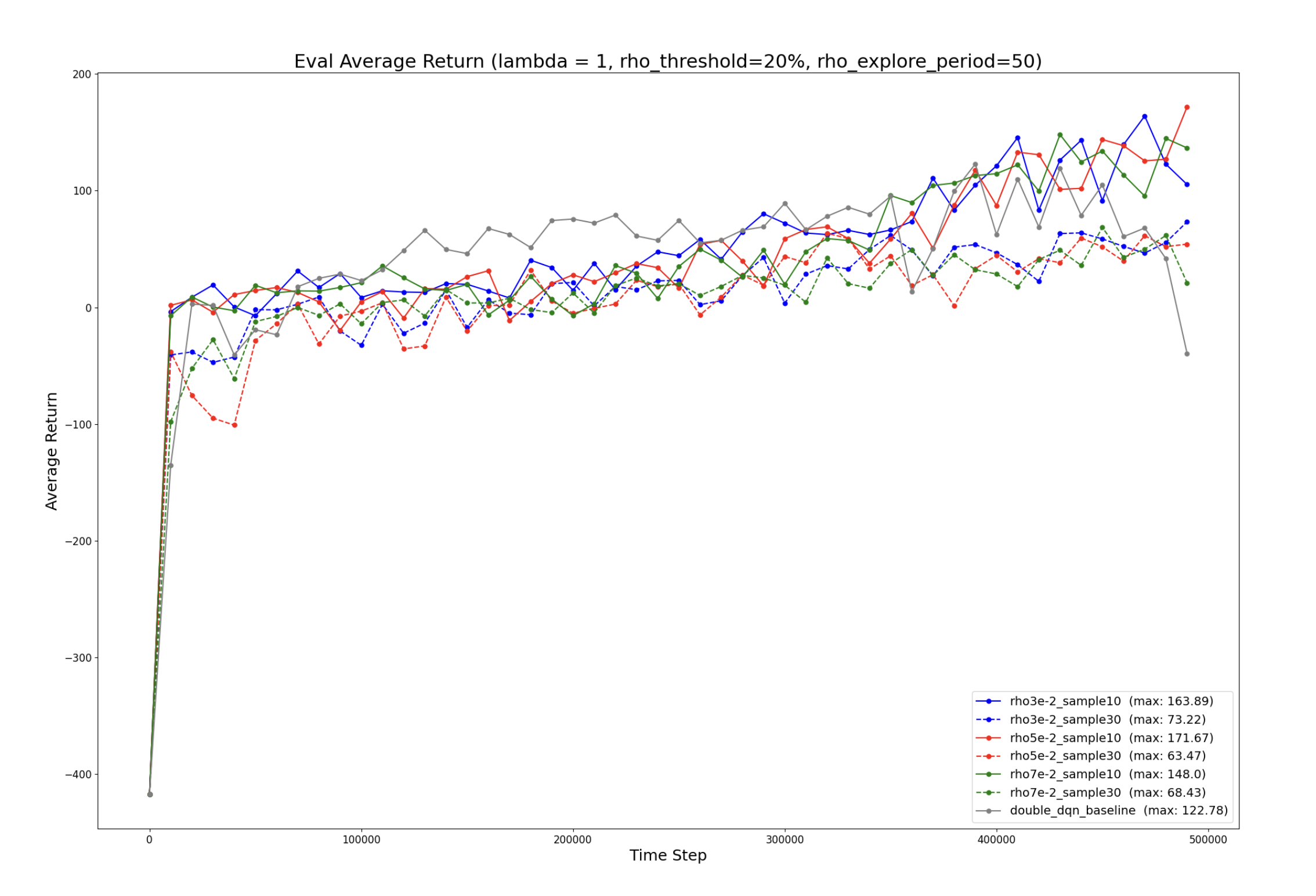}
\includegraphics[width=11cm]{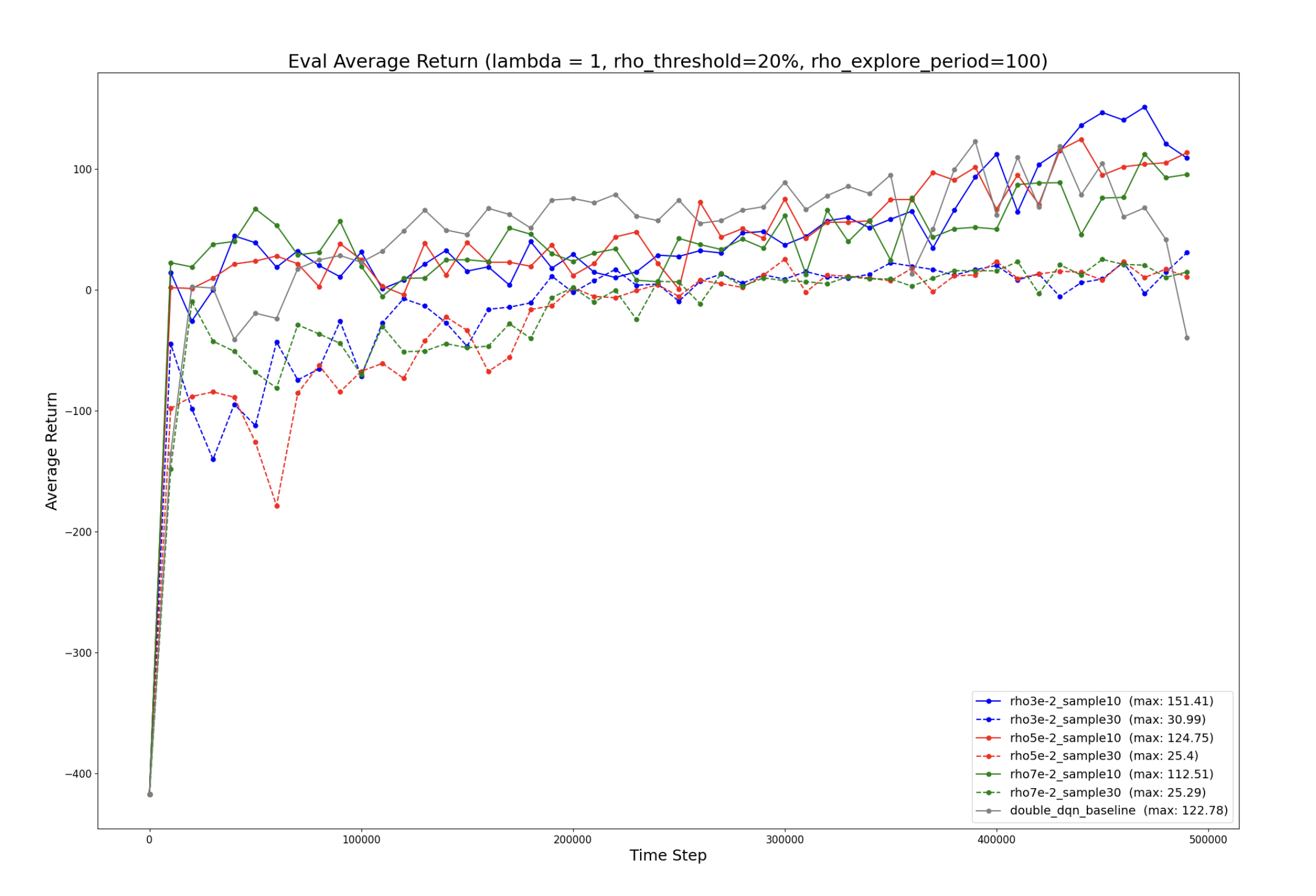}
\centering
\caption{Eval Average Return with $\lambda=1$ step; $\rho$-explore threshold=20; $\rho$-explore period=50, 100}
\label{fig:intermittent_thres20_period50_100}
\end{figure}

Results after applying "intermittent scheduling" are presented in Figures \ref{fig:intermittent_interval10} and \ref{fig:intermittent_thres20_period50_100}. Fig \ref{fig:intermittent_interval10} clearly shows that our method, in its optimal set of hyperparameters, outperforms the vanilla baseline by 49.8\% in terms of Eval Average Return. A similar trend can also be observed in Figure \ref{fig:intermittent_thres20_period50_100}, albeit by a smaller improvement margin. Furthermore, it is worth noting that almost all $\rho$-explore runs in these 2 plots are more stable than the vanilla Double DQN baseline, which oscillates significantly. We believe that this can be taken as evidence for our aforementioned hypothesis that applying $\rho$-explore to states/observations can be an effective regularizer for exploration.

\subsection{Change based exploration}

\subsubsection{Algorithm}

At the beginning of training, an RL agent knows very little about its environment. A policy is not well optimized and exploration is generally more emphasized than it is in late-stage training. Intuitively, a good exploration method may be that of choosing an action that, on average, leads to diverse states. 

To encode this idea, we bias towards actions that lead to large changes in state. We can do this by finding an action where our policy results in a state $s'$, defined as $s' = \pi(s, a)$, such that  $||s' - s||_2$ is large. We'd like to ensure that an action in question does not only lead to a next state greatly different from an original state in only very particular situations, so we consider this action over multiple initial states as an average.  We can apply this notion in both a temporally sensitive and non-sensitive way.

In a non-temporally sensitive way, we simply take a random sample from the replay buffer to attain $n$ tuples of the form $(state, action, next state)$. To find the average change, we count the number of instances for each unique action in our tuples, let the list of unique actions be $A$ and let the list of counts for each action by $C_A$. We define some $\kappa_a$ where $\kappa_a  = sum(||s_i' - s_i||_2)/ C_{Aa}$ for unique action a. Then, we normalize our list of $\kappa$s and find an action at random according to the probability distribution formed by the list of $\kappa$s. We then choose an initial state, at random, associated with this action from the initial $n$ tuples. Finally, we pass this state through our current policy to choose the action this exploration algorithm will return. 

In a temporally sensitive way, we run the above procedure with the modification that our $n$ tuples are the latest $n$ tuples added to the replay buffer. 

The hyper-parameters of this method are $\epsilon$ and $n$.

\subsubsection{Implementation Details}
Similar to $\rho$-explore, we applied change based explore along with $\epsilon$-greedy to encourage both local and global (uniformly random) exploration. What we realized in implementation is change based exploration works better in later training stages, so for observability measures, we switched from $\epsilon$-greedy to change-based exploration after some timesteps in training. 

For our exploration schedule, we adopted a linear decay schedule. We've tested with different exploration schedules, and what we've realized with ad-hoc implementations was by preventing decay from reaching zero, we're able to continue to explore nearby states in later stages that yield high returns.

For our Change Based Exploration, we have two modes in how we select the unique action that biases new states more, with $weighted$ sampling mode where the associated action is sampled from a weighted distribution given by $\kappa$ and $max$ sampling mode where we select the action with highest $\kappa$ value. From the action selected, we randomly select one of the observations in the list of observations with the same unique action and apply our policy to get our new action.

\subsubsection{Experiments}
\begin{figure}[!ht]
\includegraphics[width=11cm]{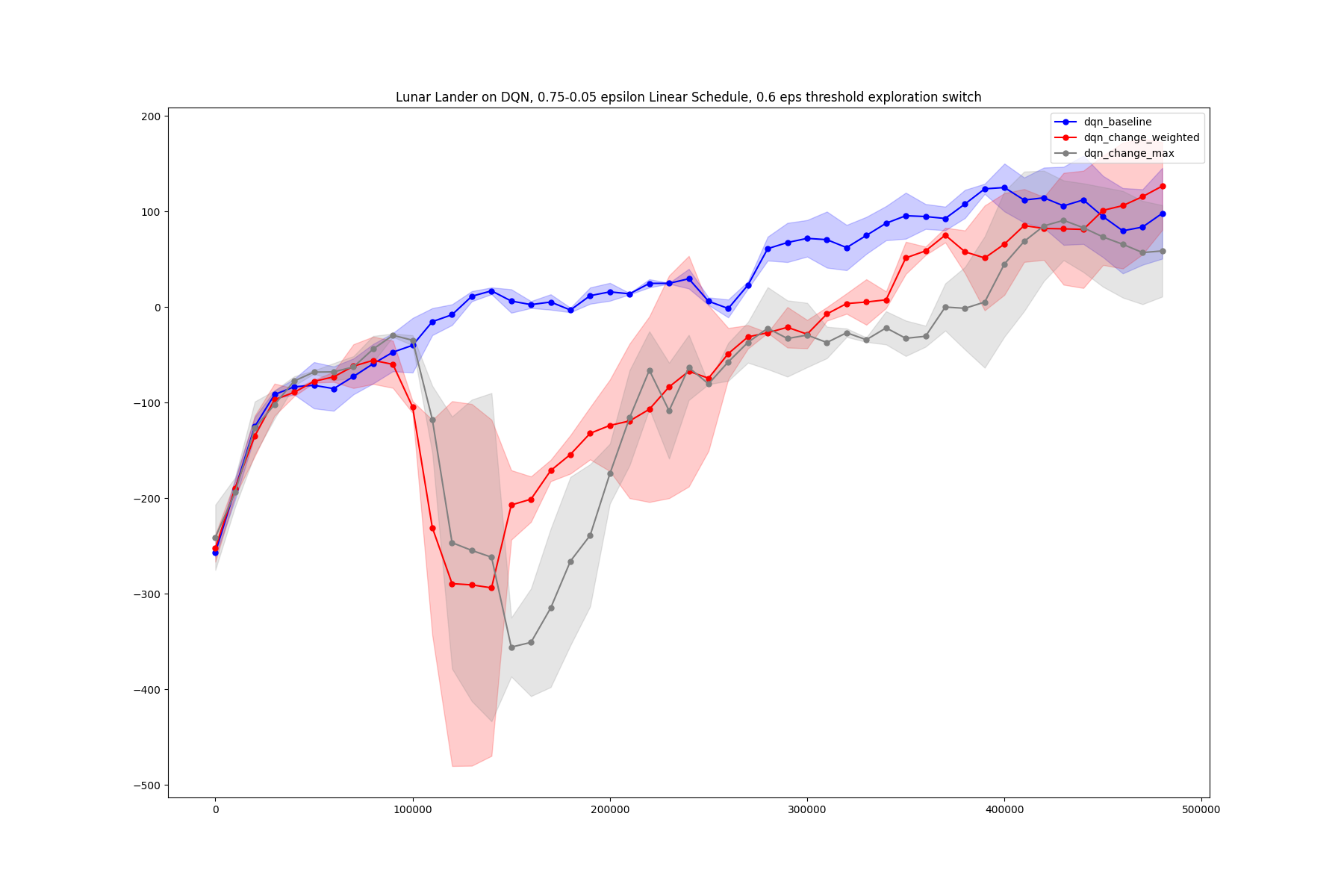}
\centering
\caption{DQN Evaluation Average Return with $\epsilon = 0.75$}
\label{fig:dqn_CB}
\end{figure}

\begin{figure}[!ht]
\includegraphics[width=11cm]{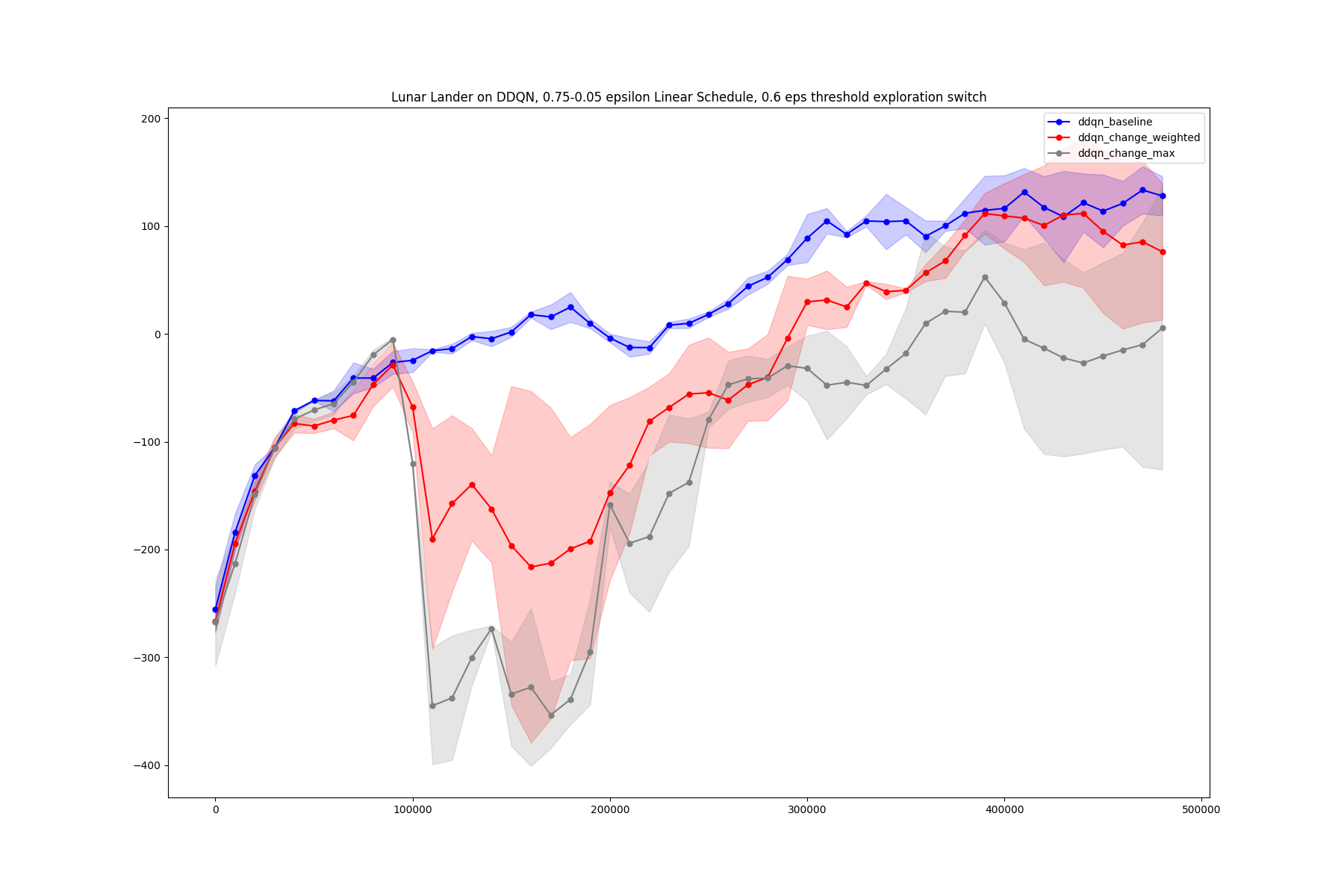}
\centering
\caption{Double DQN Evaluation Average Return with $\epsilon = 5$}
\label{fig:ddqn_CB}
\end{figure}

We experimented with both DQN and Double-DQN methods on the Lunar-Lander environment, evaluating the performances for both $weighted$ and $max$ sampling modes, with results presented in Figure \ref{fig:dqn_CB} and Figure \ref{fig:ddqn_CB}, respectively. Hyper-parameters are outlined in Table \ref{tab:DQN_CBE_HP}.

What we observe from our experiment results is that $weighted$ sampling mode has similar performances as our baseline vanilla implementations for both DQN and DDQN model, but $\max$ sampling mode across both seems to perform slightly worse. Something interesting to note from the results is the drop in returns as we switched from $\epsilon$-greedy to change based exploration, but it catches up very quickly and seems to stabilize more in later stages of training.

\begin{table}[ht]
    \centering
    \begin{tabular}{c|c}
         $\epsilon$ & .03, .05, .07, 1  \\ \hline
         $n$ & 10, 20, 40
    \end{tabular}
    \caption{DQN Change Based Exploration hyper-parameters}
    \label{tab:DQN_CBE_HP}
\end{table}

\section{Future Work}

Given more time, we would have performed grid search on $\epsilon$. This effectively manipulates the $p$-norm ball our methods considers. We would have liked to explore a probability based approach that chooses a large or small $\epsilon$ according to some probability distribution every step or a scheduled approach that reduces the size of $\epsilon$ over time. 

We also would have applied our second algorithm with local sensitivity. In a local sense, we could again rely on the notion of 'nearby' as defined by states $s'$ such that $||s' - s||_2 < \epsilon$. We could generate $n$ new states all within $\epsilon$ of the original state $s$ and generate tuples $(s, a, s')$ by applying our policy such that $s' = \pi(s, a)$. We could then follow the same procedure from algorithm 2 to calculate $\kappa_a$ for each unique action $a$ and choose an action. 

We also believe there is interesting work to be done surrounding perturbing the policy itself rather than actions or states/observations. Previous work on the topic indicates that this is quite environment dependent and potentially even initialization dependent, but there may be some consistent methodology where perturbing a particular layer under some $\epsilon$ bound results in generally improved exploration \citep{plappert2018parameter}. 

While our motivating example included an Atari environment, we did not get to try our methods beyond the Lunar Lander environment in Open AI Gym. Applying our method to Atari games or $procgen$ would be interesting. 

Another method we would have liked to try out is scoring mini-rollouts on density/frequency. We would do this by creating an exploration factor equal to the average reward of an action multiplied by its probability of occurring in a random sample from the replay buffer. 

Lastly, we would have liked to explore how we can adapt our method into a form that was more applicable to real-world use cases. Currently, we require an environment that we can step into multiple times and reset. The real world often does not offer that kind of environment. We think there may be some way to adapt our model into a model-based pre-training system where we pre-train an oracle dynamics model as a starting point for a real world RL use case. 

\section{conclusion}

The exploitation-exploration trade-off that characterizes reinforcement learning is a deeply studied problem with no one solution. Through the intuition that nearby states can inform current states during early training, we devise exploration algorithms that survey states up to an $\epsilon$, $l2$-norm distance around a state to more intelligently choose an exploratory action.  We find that our methods consistently outperform the Lunar Lander baseline and result in less variance among different runs. 

As the efficacy of exploration algorithms are generally tied to the environments they are implemented in, we believe our methods are useful in discrete-action, continuous-state environments. 

We hope that this work inspires future work in utilizing random, bounded perturbations to consider nearby states when generating exploratory actions. 

We include our code for reproducability here: \url{https://github.com/Curiouskid0423/rho_exploration}.

\bibliography{main}
\end{document}